\documentclass[journal,twoside,web]{ieeecolor}
\usepackage{generic}
\usepackage{cite}
\usepackage{amsmath,amssymb,amsfonts}
\usepackage{algorithmic}
\usepackage{graphicx}
\usepackage{algorithm,algorithmic}
\usepackage{hyperref}
\hypersetup{hidelinks=true}
\usepackage{textcomp}
\usepackage{booktabs}
\usepackage{multirow}
\usepackage[table]{xcolor}
\providecommand{\refname}{References}
\def\BibTeX{{\rm B\kern-.05em{\sc i\kern-.025em b}\kern-.08em
    T\kern-.1667em\lower.7ex\hbox{E}\kern-.125emX}}
\begin{document}
\title{Geometry-Consistent Endoscopic Representations for Image-Guided Navigation via Structured Foundation Model Adaptation}
\author{Hongchao Shu, Roger D. Soberanis-Mukul, Hao Ding, Morgan Ringel, Mali Shen, Saif Iftekar Sayed, Hedyeh Rafii-Tari, Mathias Unberath \IEEEmembership{Member, IEEE}
\thanks{Manuscript submitted 15 Apr 2026. This work is partially sponsored by Auris Health, Inc. part of Johnson \& Johnson MedTech.}
\thanks{Hongchao Shu and Roger D. Soberanis-Mukul are with the Department of Computer
Science, Johns Hopkins University, Baltimore, MD 21211 USA. (email: hshu4@jhu.edu; rsobera1@jhu.edu)}
\thanks{Hao Ding was with the Department of Computer
Science, Johns Hopkins University, Baltimore, MD 21211 USA. He is 
now with the Semaphor Surgical (email: hao@semaphorsurgical.com).}
\thanks{Morgan Ringel is now with Johnson \& Johnson MedTech, Santa Clara, CA 95054 USA (email: mringel@its.jnj.com).}
\thanks{Mali Shen is now with Johnson \& Johnson MedTech, Santa Clara, CA 95054 USA (email: mshen2@its.jnj.com).}
\thanks{Saif Iftekar Sayed is now with Johnson \& Johnson MedTech, Santa Clara, CA 95054 USA (email: ssayed3@its.jnj.com).}
\thanks{Hedyeh Rafii-Tari is now with Johnson \& Johnson MedTech, Santa Clara, CA 95054 USA (email: hrafiida@its.jnj.com).}
\thanks{Mathias Unberath is with the Department of Computer
Science, Johns Hopkins University, Baltimore, MD 21211 USA. (email: unberath@jhu.edu)}}

\maketitle

\begin{abstract}
Accurate vision-based navigation in monocular endoscopy is difficult due to limited depth cues, weak tissue texture, non-rigid deformation, and substantial appearance variation across domains, all of which complicate pose estimation, depth prediction, and image-to-anatomy alignment. Although recent vision foundation models have shown promise, their learned representations often remain insufficiently geometry-consistent, hindering stable feature correspondence and limiting their reliability for downstream navigation tasks. We propose a unified framework for learning geometry-consistent and domain-robust image representations for monocular endoscopy. The framework combines a synthetic data pipeline that provides accurate geometric supervision with Hierarchy-Aware Geometry–Semantic Adaptation, a structured alternative to standard LoRA that inserts low-rank adapters selectively across the transformer hierarchy and couples them with layer-wise training objectives to encourage geometric correspondence in intermediate features and semantic consistency in deeper features. Experiments on public and proprietary datasets show improved geometric and semantic representation quality, leading to better performance on downstream navigation tasks including pose estimation and monocular depth estimation. The learned representations show favorable synthetic-to-real transfer on clinical bronchoscopy and provide a useful initialization for adaptation to sinus endoscopy and colonoscopy under limited supervision. The framework also shows favorable scaling with model size and training data. These results support hierarchy-aware, geometry-guided adaptation as a practical approach for endoscopic representation learning.
\end{abstract}

\begin{IEEEkeywords}
surgical navigation, Endoscopic Localization, Vision Foundation Models, Representation Learning, Bronchoscopy
\end{IEEEkeywords}

\section{Introduction}
\label{sec:introduction}
\IEEEPARstart{M}{inimally} invasive endoscopic procedures such as bronchoscopy, sinus surgery, and colonoscopy reduce patient trauma and recovery time by avoiding large incisions and enabling intervention through natural orifices or small access ports. Yet these benefits come with a core limitation: clinicians must navigate complex, branching anatomy using primarily a monocular 2D image stream with limited depth cues. This makes spatial reasoning difficult, increases cognitive load, and can lead to navigation uncertainty during image-guided tasks. These challenges have motivated growing interest in computer-assisted systems for navigation support and intra-procedural decision-making. Within such systems, AI is increasingly used to extract actionable information from endoscopic images and support downstream reasoning. Representation learning is particularly important in this context because it can transform raw endoscopic observations into features that better encode geometry, viewpoint, and anatomical context. Recent progress on world models and structured visual representation learning suggests that compact latent features can support robust downstream reasoning~\cite{ha2018world,hafner2019dream}. In this work, we focus on one foundational component of that agenda for endoscopy: learning geometry-consistent image representations from individual frames. 

Learning geometry-consistent endoscopic representations from image frames is technically difficult. Endoscopic imagery is often texture-scarce and repetitive, contains strong specular highlights from fluids and tissue interfaces, and is affected by non-rigid deformation and instrument interaction. These factors break common assumptions behind geometric estimation and correspondence, making learned geometry unstable and difficult to generalize across patients, devices, and procedures~\cite{maier2013optical}. The challenge is compounded by data constraints: large-scale labeled datasets are difficult to assemble due to privacy and policy restrictions, and the most informative supervision signals such as accurate camera pose, intraoperative 3D structure, or dense depth are impractical or impossible to obtain routinely in clinical settings. As a result, methods that rely heavily on precise geometric labels or dense annotations face significant barriers to scaling and broad deployment. In addition, achieving robust generalization across clinical environments remains a persistent challenge. Endoscopic data can vary significantly across devices, imaging modalities, and hospitals due to differences in camera optics, illumination settings, and procedural protocols, causing models trained in one setting to degrade when deployed in another. These combined factors make it difficult to learn representations that simultaneously capture reliable geometric structure, remain robust under domain shifts, and scale to real-world clinical data.

Recent progress in vision foundation models has opened new opportunities for endoscopic imaging, including both adaptation of general-purpose visual backbones~\cite{oquab2023dinov2,ravi2024sam,yue2024surgicalsam,zhu2024medical} and pretraining of endoscopy-specific models~\cite{batic2024endovit,dermyer2025endodino}. Yet most existing methods are developed mainly for semantic perception and do not explicitly enforce geometric consistency in the learned representations. As a result, they remain less reliable for spatial reasoning tasks such as navigation, pose refinement, and depth understanding, where both geometric stability and cross-domain robustness are critical.

In this study, we propose a unified framework for learning geometry-consistent and domain-invariant endoscopic representations from image frames. The framework is designed to tackle three key limitations simultaneously: the lack of reliable geometric supervision, the difficulty of adapting large vision foundation models to endoscopic scenes, and the need for representations that remain robust across domains. We construct a purpose-built synthetic dataset that provides accurate geometric signals while exposing the model to diverse rendering conditions, enabling supervision that encourages domain-invariant geometric understanding without relying on scarce clinical annotations. We introduce Hierarchy-Aware Geometry–Semantic Adaptation (HGSA), which selectively inserts low-rank adapters~\cite{hu2022lora} into specific layers of a pretrained vision transformer. This structured placement is designed to better align adaptation with the transformer hierarchy, encouraging a balance between mid-level geometric feature learning and higher-level semantic abstraction while preserving the general visual knowledge of the foundation model. We employ hierarchy-aware layer-wise objectives that combine feature-warping–based geometric supervision with global semantic alignment. By enforcing geometric correspondence at intermediate feature levels and semantic consistency at deeper layers, the framework encourages representations that maintain spatial structure while remaining robust across visual domains. Together, these components form an end-to-end pipeline for adapting a generic vision foundation model into a geometry-consistent representation that supports downstream image-guided navigation tasks in endoscopic environments.

Through a set of experiments, we evaluate the representation quality, practical utility, and generalization of the proposed framework. Controlled studies first examine the effectiveness of the HGSA adaptation strategy through linear probing tasks that separately assess semantic and geometric representation quality, using scene classification and depth estimation as complementary indicators. The practical value of the learned representations is then validated in clinically relevant image-guided navigation settings, where improved geometric consistency leads to more stable pose optimization and depth prediction. The robustness and scalability of the framework are evaluated through synthetic-to-real transfer on bronchoscopy data, where the encoder is trained solely on synthetic data but applied to real patient videos. We further investigate cross-domain transfer by adapting the bronchoscopy-trained encoder to sinus and colonoscopy datasets under limited target-domain supervision, while also analyzing performance trends across larger backbone models and increasing training data.

The contribution of this work lies in the following perspectives:
\begin{itemize}
    \item We introduce a unified pipeline that integrates a purpose-built dataset, the HGSA mechanism and hierarchy-aware, layer-wise objectives that jointly enforce geometric consistency and semantic robustness to learn endoscopic representations.
    \item We show through controlled experiments that HGSA is an effective adaptation strategy for improving representation quality, and that the resulting representations remain useful across domains while supporting downstream image-guided navigation tasks including pose estimation and depth estimation.
    \item We demonstrate strong sim-to-real generalization, where representations trained on synthetic bronchoscopy data transfer effectively to real patient videos, and further study cross-domain transfer to sinus and colonoscopy datasets under limited supervision, together with scalability analyses across larger transformer backbones and increasing training data.
\end{itemize}

\begin{figure*}[!t]
\centerline{\includegraphics[width=\textwidth]{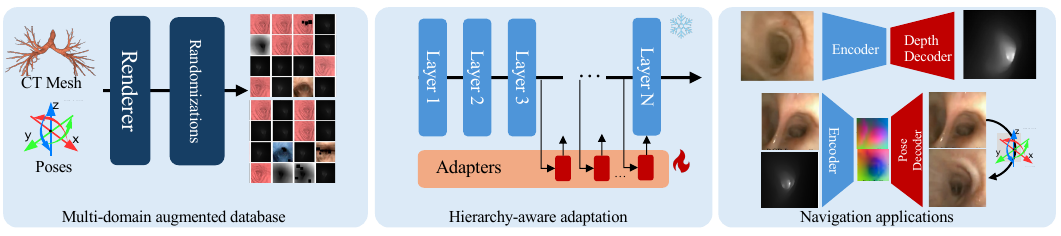}}
\caption{Overview of the proposed framework.
The figure shows an example instantiation in bronchoscopy. Synthetic 3D anatomical models are rendered under sampled camera poses and augmented to form a multi-domain training set. The HGSA inserts lightweight adapters into selected transformer layers to learn geometry-consistent and domain-robust representations, which are then used in downstream navigation-related tasks such as depth estimation and pose optimization.}
\label{fig:overview}
\end{figure*}

\section{Related Works}
\subsection{Vision Foundation Models for Endoscopic Imaging}
Foundation models (FMs) are increasingly used in medical imaging because large-scale self-supervised pre-training can reduce reliance on densely annotated datasets~\cite{phuntsho2025adaptationfoundationmodelsmedical}. 
In endoscopy, early work mainly adapted general vision foundation models such as DINO~\cite{oquab2023dinov2} and SAM~\cite{ravi2024sam} through zero-shot transfer or parameter-efficient tuning. 
Representative examples include SAM-based medical adaptations~\cite{zhu2024medical,yue2024surgicalsam,shen2024fastsam3d,soberanis2024gsam+,killeen2025fluorosam} and adapter-based methods such as ST-Adapter~\cite{pan2022st}. 
Although these approaches improve transfer with limited trainable parameters, they remain constrained by the domain gap between natural images and endoscopic data, often lacking robustness to low-contrast tissue appearance and complex procedural dynamics~\cite{yue2024surgicalsam}.

\subsection{Domain-Specific Endoscopic Foundation Models}
To better address this gap, several works have developed endoscopy-specific foundation models trained directly on large surgical datasets. 
Examples include image-based models such as EndoViT~\cite{batic2024endovit} and EndoDINO~\cite{dermyer2025endodino}, as well as video-oriented models such as Endo-FM~\cite{wang2023foundation} and EndoMamba~\cite{tian2025endomamba}. 
These models capture endoscopic appearance and motion more effectively than general-purpose backbones, but require large curated datasets and still remain sensitive to cross-domain variation such as differences in hardware, illumination, and procedure type. 
Moreover, they do not explicitly enforce 3D geometric consistency~\cite{dermyer2025endodino}.

\subsection{Domain-Invariant and Geometry-Aware Representation Learning}
Robust endoscopic representations should ideally be both domain-invariant and geometry-consistent, yet existing work often addresses these goals separately. 
Domain generalization methods, including adversarial feature alignment~\cite{li2025monocular}, federated optimization~\cite{qin2025enhancing}, and simulation-based appearance diversification through realistic rendering and domain randomization~\cite{gao2023fully,gao2023synthetic}, mainly improve robustness to visual appearance shifts without explicitly enforcing 3D spatial structure. 
Conversely, geometry-focused approaches, such as Surgical-DINO~\cite{cui2024surgical} and foundation-model-based methods combined with NeRF, improve depth or multi-view consistency but are often more vulnerable to real clinical appearance shifts~\cite{deng2025best}. 
More recent work has begun to connect these directions. For example, BronchOpt~\cite{shu2025bronchopt} uses a fine-tuned foundation model within an iterative pose optimization framework for bronchoscopy navigation. 
In contrast, our work focuses on learning geometry-consistent and domain-robust representations directly through structured adaptation and synthetic geometric supervision.

\section{Methodology}
\subsection{Framework Overview}
Fig.~\ref{fig:overview} illustrates the proposed framework, which consists of three stages. 
A multi-domain augmented training database (Sec. \ref{sec:database}) is constructed by rendering paired endoscopic views from anatomical 3D models under sampled camera poses, together with appearance augmentation that increases visual diversity while preserving geometric consistency. 
These training pairs are then used to adapt a pretrained vision transformer through the proposed Hierarchy-Aware Geometry--Semantic Adaptation (HGSA) module (Sec. \ref{sec:hgsa}), which selectively inserts low-rank adapters into layers of the backbone and applies layer-wise supervision (Sec. \ref{sec:supervision}) to encourage geometric correspondence in intermediate features and semantic consistency in deeper features. 
The adapted encoder is finally evaluated in downstream navigation-related tasks (Sec. \ref{sec:app}), including monocular depth estimation and endoscope pose estimation, to assess the utility of the learned representation.

The following subsections describe each component in detail.

\subsection{Latent Representation Modeling of the Endoscopic Environment}\label{sec:modeling}
We seek to learn a visual representation that captures the structural properties of the endoscopic scene from individual image frames. Given an input image $I \in \mathbb{R}^{H \times W \times 3}$, a vision foundation model $f_\theta(\cdot)$ produces a latent feature map
\begin{align}
    z = f_\theta(I), \quad z \in \mathbb{R}^{N \times C},
\end{align}
where $N$ is the number of spatial tokens and $C$ is the feature dimension. The desired representation should preserve scene geometry, remain robust across imaging domains, and maintain semantic separability.

Geometric consistency requires that spatial relationships between regions are preserved in the latent space. For two tokens $z_i$ and $z_j$, their similarity is defined as
\begin{align}
    s_{ij} = \frac{z_i^\top z_j}{\|z_i\|\|z_j\|},
\end{align}
which should remain stable for regions corresponding to the same underlying anatomical structure.

Domain invariance requires the representation to generalize across imaging conditions. Let $I_s \sim \mathcal{D}_s$ and $I_r \sim \mathcal{D}_r$ denote synthetic and real clinical images, respectively. For structurally similar scenes, the representation should remain consistent across domains:
\begin{align}
    f_\theta(I_s) \approx f_\theta(I_r).
\end{align}

Semantic discriminability requires features from different anatomical categories to remain separable in the latent space.


\subsection{Hierarchy-aware Geometry-Semantic Adaptation}\label{sec:hgsa}
Vision foundation models pretrained on large-scale natural image datasets provide strong general-purpose visual representations, but their internal feature hierarchy is not optimized for the geometric reasoning required in endoscopic environments. In particular, different transformer layers capture different levels of abstraction: intermediate layers typically encode spatial structures and local correspondences, while deeper layers capture higher-level semantic context. This observation motivates a hierarchy-aware adaptation strategy, where parameter-efficient adapters are inserted at selected transformer layers to better align adaptation with the geometric and semantic roles of different transformer stages.

Let $f_{\theta_0}$ denote a pretrained vision transformer with L layers. HGSA adapts the model by inserting lightweight adapters into a subset of layers $\mathcal{L}_a \subset \{1,\ldots,L\}$, producing an adapted encoder
$z = f_{\theta_0,\phi}(I)$
where $\phi$ denotes the trainable adapter parameters. Rather than uniformly adapting all layers, HGSA determines the adapter configuration through a coarse-to-fine hierarchical search over the design space
\begin{align}
    \mathcal{C} = \{ (\mathcal{L}_a, r, \alpha, M) \}  
\end{align}
where $\mathcal{L}_a$ denotes adapter layer placement, $r$ the adapter rank, $\alpha$ the scaling factor, and $M$ the targeted attention projection modules. Exhaustively exploring this space would be computationally prohibitive, therefore HGSA adopts a hierarchical coarse-to-fine search strategy that progressively refines the configuration space. The search prioritizes factors with larger influence on representation behavior, beginning with adapter placement, followed by projection targets, and finally adapter capacity parameters. The search proceeds in three stages while retaining only the strongest candidates at each step. First, adapter insertion patterns along the transformer hierarchy are explored to determine the most effective layer subset $\mathcal{L}_a$. Guided by the hierarchical nature of transformer representations~\cite{dosovitskiy2021imageworth16x16words,caron2021emerging}, candidate placements include early, middle, late, and several late-layer combinations. Next, HGSA evaluates different attention projection targets $M$, inserting adapters into combined projections including $V$, $QV$, $QVK$, and $QVKO$. Finally, the search refines adapter capacity through rank–scaling parameterization, evaluating rank–scaling pairs $(r,\alpha)$, configurations commonly used to maintain stable adaptation strength~\cite{hu2022lora,zhang2023adalora}. 

Each candidate configuration $c \in \mathcal{C}$ defines an adapted encoder $f_{\theta_0,\phi}^{(c)}$, which is evaluated using linear probing on two complementary tasks reflecting semantic and geometric representation quality. 



\subsection{Geometry and Semantic-aware Supervision}\label{sec:supervision}
\begin{figure}[!t]
\centerline{\includegraphics[width=\columnwidth]{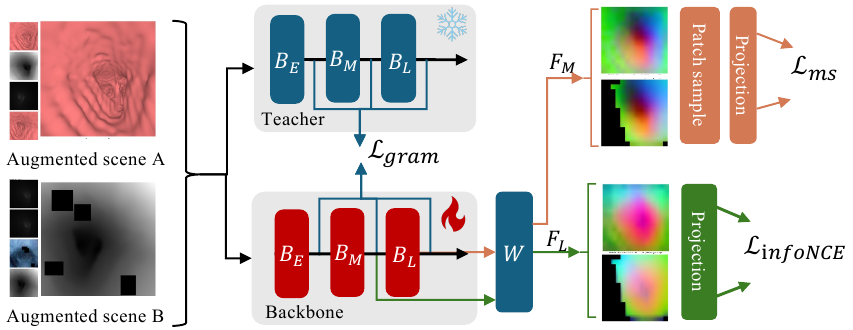}}
\caption{Overview of the proposed geometry–semantic representation learning pipeline.
Synthetic endoscopic images with known geometry are used to supervise representation learning. Disparity (or depth) is estimated from real images, while rendered depth from synthetic views provides geometric grounding. The encoder is trained to align features across domains under geometric constraints, encouraging both domain invariance and correspondence consistency.}
\label{fig:train}
\end{figure}
To learn representations that preserve spatial structure while maintaining semantic robustness, we design a supervision framework that jointly enforces geometry-aware local correspondence and semantic-level instance consistency. The supervision operates on two complementary representations extracted from the encoder: spatial tokens that retain spatial detail and global tokens that summarize scene-level semantics.
\subsubsection{Geometry-aware Feature Correspondence}
Given two images $I_a$ and $I_b$ observing the same scene from different viewpoints, the encoder produces spatial token features
$z_a = f_{\theta}(I_a), \qquad z_b = f_{\theta}(I_b)$,
where $z_a,z_b \in \mathbb{R}^{N \times C}$. Using rendered depth and relative camera pose, we compute dense pixel correspondences between the two views, resulting in a ground-truth flow field $F_{b\rightarrow a}$. The feature map of view $I_b$ is then warped into the coordinate frame of $I_a$:
\begin{align}
\tilde{z}_b = \mathcal{W}(z_b, F_{b\rightarrow a})
\end{align}
where $\mathcal{W}(\cdot)$ denotes a differentiable feature warping operator. A valid overlap mask $M$ identifies spatial regions visible in both views. Geometry-aware supervision is applied through a multi-scale dense token loss \begin{align}
    \mathcal{L}_{ms} = \sum_{l \in \mathcal{L}}w_{\text{mask}}\left(\mathcal{L}_{\text{PatchNCE}}^{(l)} +\mathcal{L}_{\text{cos}}^{(l)}\right)
\end{align}
where $w_{\text{mask}}$ denotes the valid patch mask weights derived from geometric visibility, $\mathcal{L}_{\text{PatchNCE}}$ is the patch-level contrastive loss~\cite{park2020contrastive} on geometry-aligned tokens, and $\mathcal{L}_{\text{cos}}$ is the cosine reprojection loss between warped and reference features. This loss is applied to mid-level transformer features, where spatial structure is most strongly preserved.

\subsubsection{Semantic-aware Representation Regularization}
While spatial tokens capture local spatial structure, we also construct global tokens that summarize the entire frame. These are obtained by masked pooling of dense features followed by projection heads, producing compact view-level embeddings $g_a = h(z_a), \qquad g_b = h(z_b)$.
Unlike the spatial tokens, these global representations capture scene-level semantics rather than spatial detail. Semantic consistency is enforced using an instance-level contrastive objective $\mathcal{L}_{\text{InfoNCE}}$~\cite{oord2018representation}, which aligns global features extracted from paired views. Only view pairs with valid geometric overlap contribute to the loss. This encourages different observations of the same scene to produce similar global representations while separating unrelated scenes. In the transformer hierarchy, this loss is applied to late-layer features, where semantic abstraction is strongest.

\subsubsection{Semantic Structure Preservation via Teacher Distillation}
During training, strong global contrastive supervision may encourage the model to focus excessively on scene-level semantics, potentially degrading the spatial fidelity of patch representations. To prevent this effect, we incorporate a Gram-based feature regularization $\mathcal{L}_{\text{Gram}}$~\cite{simeoni2025dinov3} using a frozen DINO teacher model~\cite{oquab2023dinov2}. This loss matches the correlation structure of student and teacher patch features across selected layers. By preserving the relational structure among patch embeddings, this regularization helps maintain spatial resolution in the learned representation while allowing semantic adaptation.

The final training objective combines geometry-aware dense supervision with semantic-level regularization:
\begin{align}
    \mathcal{L} = \lambda_{ms}\mathcal{L}_{ms} + \lambda_{nce}\mathcal{L}_{\text{InfoNCE}} + \lambda_{gram}\mathcal{L}_{\text{gram}}
\end{align}
where the weighting coefficients $\lambda_{ms}$, $\lambda_{nce}$, and $\lambda_{gram}$ balance the contributions of the three supervision signals.

\subsection{Physically Grounded Dataset with Multi-Level Domain Randomization}\label{sec:database}
To provide reliable geometric supervision without relying on clinical annotations, we construct a geometry-supervised synthetic dataset with domain randomization. The dataset is generated from CT-derived anatomical models by rendering virtual endoscopic views along sampled camera trajectories. For each pose, RGB images and depth maps are rendered. To increase appearance diversity, rendered RGB images are translated to realistic using trained CycleGAN~\cite{zhu2017unpaired}. In addition, disparity predictions generated by an off-the-shelf network~\cite{paruchuri2024leveraging} are included as an alternative modality. Overall there are 4 different image modalities.

During training, input modality randomization is applied for each view. The encoder input is randomly selected from all modalities. Additional augmentations are applied to both RGB and depth inputs. RGB images undergo photometric transformations, blur and noise perturbations, thin-plate spline deformation, and coarse dropout. Depth maps are augmented with scale perturbations, depth-to-disparity conversion, blur, noise, and dropout. Training also includes pair-level randomization, where image pairs are sampled from precomputed cross-view relationships and reshuffled each epoch. For geometry supervision, valid patch correspondences are randomly sampled from overlapping regions between paired views.

\subsection{Applications to Image-guided Navigation}\label{sec:app}
we apply the trained encoder to two core tasks in image-guided surgical navigation: vision-based endoscope pose estimation$ \hat{T} = h_{pose}(z, I), \quad \hat{T} \in SE(3)$ where $T$ denotes the 6-DoF camera pose.
And monocular endoscope depth estimation $\hat{D} = h_{depth}(z), \quad \hat{D} \in \mathbb{R}^{H \times W}$, where $D$ denotes the depth map. These tasks capture two fundamental components of intraoperative spatial understanding.

In both applications, the encoder trained with the proposed framework is kept frozen and used purely as a feature extractor.

\section{Experiments Settings}
\subsection{Dataset}
We study three endoscopic modalities: bronchoscopy, sinus endoscopy, and colonoscopy. All datasets consist of monocular endoscopic video frames and are preprocessed using a shared protocol, including center-cropping and resizing to $200 \times 200$.

Synthetic training pairs are generated from 3D anatomy by rendering paired endoscopic views under sampled camera poses, followed by modality-specific appearance translation using separately trained CycleGAN models~\cite{zhu2017unpaired}. The bronchoscopy synthetic training set used for the main representation learning stage contains $481{,}650$ image pairs. For cross-procedure transfer, the sinus and colonoscopy synthetic training sets used for few-shot target-domain fine-tuning contain $8{,}944$ and $18{,}832$ image pairs, respectively.

For evaluation, the real bronchoscopy dataset contains $8$ clinical procedures and $20{,}555$ frames, but no frame-wise aligned 6-DoF bronchoscope poses, so evaluation relies on alignment-based proxy metrics. The sinus dataset contains $4$ sequences and $1{,}107$ frames with 6-DoF poses. For colonoscopy, we use a matched subset of the C3VD dataset~\cite{bobrow2023colonoscopy} consisting of $4$ sequences and $1{,}176$ frames with 6-DoF poses. Additional dataset construction and split details are provided in the supplementary material.

\subsection{Representation Probing}\label{sec:probe}
We evaluate representation quality using linear probing on two complementary tasks measuring semantic and geometric properties. For both tasks, we use cross-fold evaluation split by patient case to prevent leakage between training and validation sets. For semantic probing, we train a linear classifier on frozen encoder features to perform scene classification over seven anatomical landmark categories. Performance is measured using Accuracy (Acc) and F1-score (F1). For geometric probing, monocular depth estimation is performed by training a lightweight depth prediction head on frozen encoder features. Performance is evaluated using AbsRel, LogRMSE, and $\delta_1$~\cite{eigen2014depth}. To compare encoder configurations, we compute a combined score
\begin{align}
    S_{probe} = \lambda_{1} S_{geo} + \lambda_{2} S_{sem}
\end{align}
where$S_{geo}=e^{-\text{AbsRel}} + e^{- \text{LogRMSE}} + \delta_1$, $S_{sem}=\text{Acc} + \text{F1}$. $\lambda_1$ and $\lambda_2$ balance contributions of geometric and semantic evaluation.

\subsection{Navigation Task Evaluation}\label{sec:metric}
We evaluate the learned representations through two downstream navigation-related tasks: vision-based endoscope pose estimation and monocular endoscope depth estimation. In both applications, the encoder is kept frozen and used purely as a feature extractor within established downstream pipelines, allowing the effect of representation quality to be assessed independently of task-specific architecture changes.

We intentionally adopt established downstream architectures to isolate the contribution of the learned representation and avoid conflating encoder improvements with changes in task-specific network design.

\subsubsection{Endoscope pose estimation}
Pose estimation follows the framework design of BronchOpt~\cite{shu2025bronchopt}, where the frozen encoder provides visual features for pose regression. Performance is evaluated using the combined pose error consisting of $\mathcal{L}_{2}$ translation distance and geodesic rotation distance $\mathcal{L}_{geo}$. In addition, we employ image similarity–based metrics~\cite{shu2025bronchopt} as proxy measures of pose accuracy by evaluating the alignment between rendered and observed views. Specifically, we compute Normalized Correlation (NC), Depth Similarity (DS), and Scale-Invariant Structural Alignment (SI)~\cite{eigen2014depth}, where higher similarity indicates better geometric alignment under the estimated pose.

\subsubsection{Monocular depth estimation}
Depth estimation follows the network design of DPT~\cite{ranftl2021vision}, where a lightweight decoder predicts dense depth maps from encoder features. Performance is evaluated using standard metrics including AbsRel, LogRMSE, and $\delta_1$.

Cross-domain evaluation on the sinus endoscopy and colonoscopy datasets is performed by retraining only the task-specific heads using a small amount of target-domain data while keeping the encoder frozen. Benchmark comparisons are conducted against alternative pretrained backbones under the same protocol.




\section{Results and Discussion}
\begin{figure*}[!t]
\centerline{\includegraphics[width=\textwidth]{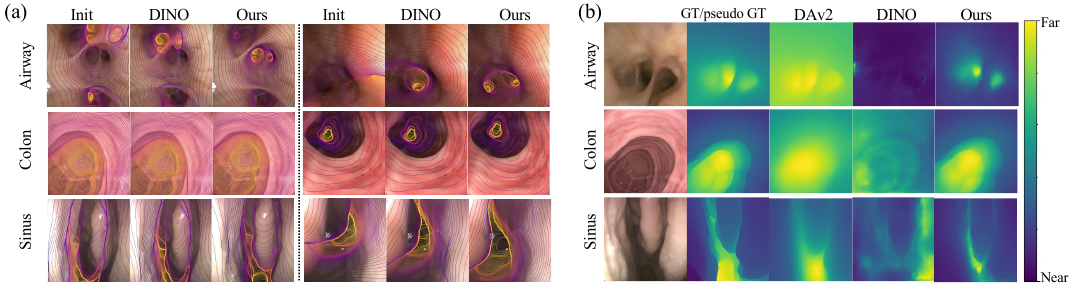}}
\caption{Qualitative results for the navigation tasks.(a) Pose estimation visualization across airway, colon, and sinus domains. Rendered depth contours are overlaid on input images for the initial pose (Init). Init shows the alignment before optimization, while DINO~\cite{simeoni2025dinov3} and Ours use the same optimization pipeline with different encoders under same training protocol. The proposed method achieves better alignment with anatomical structures.
(b) Cross-domain depth estimation. From left to right: input, ground truth (pseudo GT for Airway), DAv2~\cite{yang2024depth}, DINO~\cite{simeoni2025dinov3}, and Ours. The proposed method produces more accurate and structurally consistent depth predictions.}
\label{fig:visual}
\end{figure*}
This section presents experimental results and analysis of the proposed framework. We first evaluate the HGSA strategy through controlled ablation studies to understand how adapter placement and configuration influence representation quality. We then analyze the learned representations to assess domain alignment and geometric consistency, providing both qualitative and quantitative evidence of domain-invariant and geometry-preserving behavior. We evaluate the practical effectiveness of the learned representations in clinically relevant image-guided navigation tasks, including endoscope pose estimation and monocular depth estimation, and examine cross-domain generalization to other endoscopic procedures. 

\subsection{HGSA Configuration Study}
\begin{figure}[!t]
\centerline{\includegraphics[width=\columnwidth]{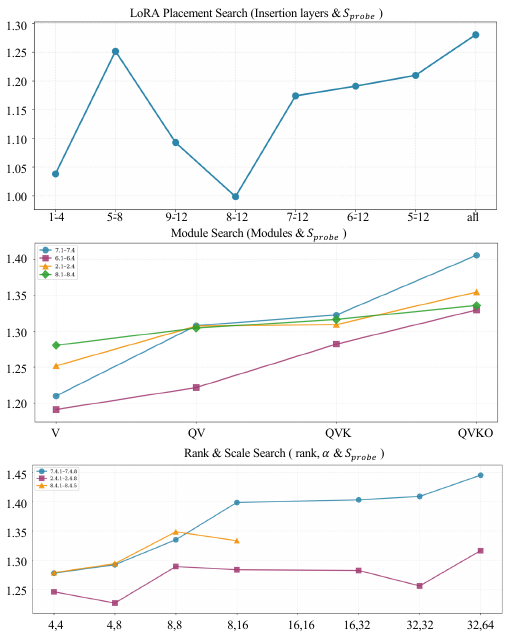}}
\caption{HGSA behavior across design stages.
Top: adapter placement across transformer layers. Middle: attention module selection. Bottom: rank–scale analysis. Scores correspond to $S_{probe}$. While early configurations show non-monotonic behavior, progressively structured HGSA design leads to stable and consistent performance gains, highlighting the importance of hierarchy-aware adaptation.}
\label{fig:HGSA}
\end{figure}

\begin{table*}[t]
\caption{Progressive HGSA configuration search over layer placement, attention-module selection, and LoRA rank/scale.
Performance is measured by the semantic score $S_{sem}$, geometric score $S_{geo}$, and their combined probing score $S_{\text{probe}}$.
Within each stage, the best, second-best, and third-best configurations are highlighted with decreasing green intensity.
The final configuration (Cfg~7.4.8) is used in subsequent experiments.
}
\centering
\small
\setlength{\tabcolsep}{5pt}

\begin{tabular}{c c c c c | c c c c c | c c c c c}
\toprule

\multicolumn{5}{c|}{Stage 1: Layer Placement Search} &
\multicolumn{5}{c|}{Stage 2: Target Module Search} &
\multicolumn{5}{c}{Stage 3: LoRA Rank \& Scale Search} \\

\cmidrule(lr){1-5}
\cmidrule(lr){6-10}
\cmidrule(lr){11-15}

Cfg & Layers & $S_{sem}$ & $S_{geo}$ & $S_{probe}$ &
Cfg & Modules & $S_{sem}$ & $S_{geo}$ & $S_{probe}$ &
Cfg & $(r,\alpha)$ & $S_{sem}$ & $S_{geo}$ & $S_{probe}$ \\

\midrule
DINO & w/o LoRA  & 0.805 & 0.456 & 1.261 & 
7.1 & V    & 0.685 & 0.519 & 1.204 &
7.4.1 & (4,4)  & 0.745 & 0.529 & 1.274 \\

1 & 1–4  & 0.525 & 0.514 & 1.039 &
7.2 & QV   & 0.765 & 0.544 & 1.309 &
7.4.2 & (4,8)  & 0.770 & 0.523 & 1.293 \\

2 & 5–8  & 0.690 & 0.562 & \cellcolor{green!50}1.252 &
7.3 & QVK  & 0.760 & 0.563 & 1.323 &
7.4.3 & (8,8)  & 0.785 & 0.551 & 1.336 \\

3 & 9–12 & 0.640 & 0.457 & 1.097 &
7.4 & QVKO & 0.830 & 0.576 & \cellcolor{green}1.406 &
7.4.4 & (8,16) & 0.820 & 0.579 & 1.399 \\

4 & 8–12 & 0.520 & 0.481 & 1.001 &
2.4 & QVKO & 0.735 & 0.624 & \cellcolor{green!50}1.359 &
7.4.5 & (16,16) & 0.830 & 0.576 & \cellcolor{green!20}1.406 \\

5 & 7–12 & 0.665 & 0.505 & 1.170 &
8.4 & QVKO & 0.785 & 0.550 & \cellcolor{green!20}1.335 &
7.4.6 & (16,32) & 0.820 & 0.583 & 1.403 \\

6 & 6–12 & 0.675 & 0.517 & 1.192 &
 &  &  &  & &
7.4.7 & (32,32) & 0.830 & 0.576 & \cellcolor{green!20}1.406 \\

7 & 5–12 & 0.680 & 0.519 & \cellcolor{green!20}1.199 &
 &  &  &  & &
\textbf{7.4.8} & (32,64) & 0.850 & 0.592 & \cellcolor{green}1.442 \\

8 & Full LoRA & 0.735 & 0.547 & \cellcolor{green}1.282 & \\

\bottomrule
\end{tabular}
\label{tab:hgsa}
\end{table*}
The results of the hierarchical adaptation search are summarized in Table~\ref{tab:hgsa}. 
The search is performed progressively over three dimensions. 
We first search LoRA layer placement along the transformer hierarchy and retain the top three configurations according to the combined probing score $S_{\text{probe}}$. For clarity, Table~\ref{tab:hgsa} reports, at each stage, the top three configurations together with the full set of results from one representative branch to show the trend within that search dimension (e.g., Cfgs.~7.1–7.4 in Stage 2). We then evaluate attention-module targets within the retained placement branches and subsequently refine LoRA rank and scaling based on the top Stage 2 configurations. The final selected model is used in subsequent experiments, and the full set of evaluated configurations ($41$ in total) is provided in the supplementary material.

Several patterns observed in this search are consistent with prior findings on transformer representations and parameter-efficient adaptation (Fig.~\ref{fig:HGSA}). 
First, performance is sensitive to adapter placement, reflecting the hierarchical structure of transformer features. 
Prior work has shown that intermediate layers tend to preserve more spatial structure, whereas deeper layers encode increasingly semantic information~\cite{caron2021emerging}. 
Consistent with this view, the strongest probing results are obtained when adaptation extends beyond the earliest transformer layers. 
Second, the rank search shows that increasing adapter rank improves performance up to a relatively stable regime. 
This trend agrees with prior studies on LoRA and related parameter-efficient tuning methods, which show that higher ranks increase adaptation capacity but eventually reach diminishing returns~\cite{hu2022lora,zhang2023adalora}.

Beyond these expected trends, the search also reveals observations that are particularly relevant to endoscopic representation learning. 
The probing results suggest a geometry--semantic trade-off: some configurations (e.g., Cfg~8) improve semantic separability while weakening geometric consistency, whereas others (e.g., Cfg~2) favor geometric accuracy at the cost of semantic discrimination. 
Within this trade-off, the HGSA search provides a practical way to identify configurations that perform well on both criteria. 
The module search further shows that adapting multiple attention projections jointly yields stronger representations than modifying individual projections in isolation, suggesting that geometry-aware correspondence learning benefits from coordinated adaptation across attention pathways. 
Moreover, although full-hierarchy adaptation achieves the highest score during the placement search, the final best configuration emerges from Cfg~7 (layers 5--12) after module and rank refinement. 
This indicates that effective adaptation is not determined by placement alone, but by the interaction among layer selection, target modules, and adaptation capacity.

Although these results support the usefulness of hierarchy-aware adaptation, they do not by themselves fully disentangle the effect of adapter placement from the broader training recipe. 
We therefore interpret HGSA primarily as an empirically effective configuration strategy rather than as a uniquely identified mechanism.

\begin{table}[t]
\caption{
Ablation study of the proposed representation learning framework. Performance is evaluated using metrics defined in Sec.~\ref{sec:probe}.
}
\centering
\small
\setlength{\tabcolsep}{16pt}

\begin{tabular}{lccc}
\toprule
Variant & $S_{sem}$ & $S_{geo}$ & $S$ \\
\midrule
w/o HGSA & 0.79 & 0.52 & 1.31 \\
w/o $\mathcal{W}arp$ &0.74 &0.52 & 1.26\\
w/o $\mathcal{L}_{ms}$ & 0.85 & 0.61 & 1.46 \\
w/o $\mathcal{L}_{InfoNCE}$ & 0.63 & 0.56 & 1.19 \\
w/o $\mathcal{L}_{Gram}$ & 0.88 & 0.60 & 1.48 \\
\midrule
\textbf{Full model} & \textbf{0.91} & \textbf{0.64} & \textbf{1.55} \\
\bottomrule
\end{tabular}
\label{tab:ablation}
\end{table}

\subsection{Component Ablation Study}
To assess the contribution of each component, we perform controlled ablations by removing one module at a time from the full model while keeping all other training settings unchanged. Representation quality is evaluated using the same probing tasks described in Sec.~\ref{sec:probe}, and the results are summarized in Table~\ref{tab:ablation}. Removing HGSA lowers both semantic and geometric scores, indicating that hierarchy-aware configuration improves the joint quality of the learned representation. Among the loss terms, removing $L_{\text{InfoNCE}}$ causes the largest drop, especially in semantic performance, suggesting that contrastive alignment is the main driver of semantic separability. Removing $L_{\text{ms}}$ produces a smaller but consistent degradation, particularly in the geometric score, indicating that multi-scale geometric supervision provides complementary support for spatial correspondence. Removing $L_{\text{Gram}}$ leads to a modest decline, suggesting an additional but non-dominant regularization benefit. Performance also decreases when warp supervision is removed, showing that explicit correspondence learning through geometric warping contributes beyond global contrastive supervision alone. Overall, the full model performs best across all three probing metrics, indicating that HGSA, contrastive alignment, geometric supervision, warp-based correspondence learning, and regularization act as complementary components of the final representation-learning framework.

\subsection{Geometry–Semantic Representation Analysis}
\begin{figure}[!t]
\centerline{\includegraphics[width=\columnwidth]{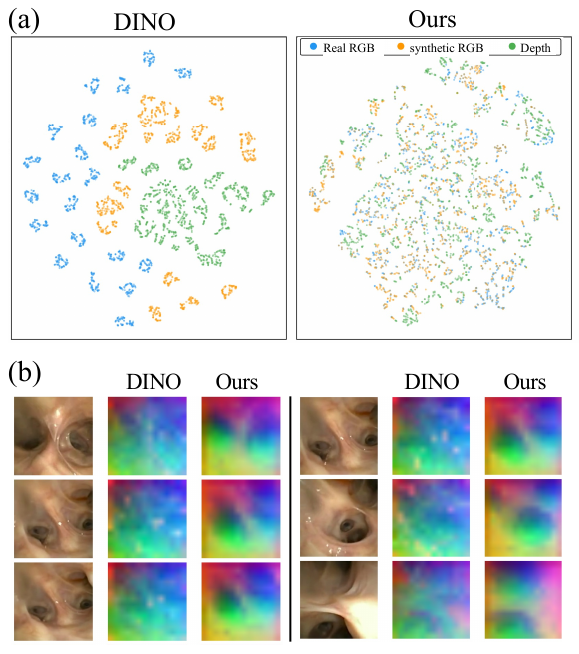}}
\caption{(a) t-SNE visualization of feature embeddings from real RGB, synthetic RGB, and depth images. The pretrained DINO encoder produces domain-separated clusters, while the proposed method aligns features from different modalities into a unified space, indicating improved domain invariance.
(b) Feature map visualization across views. Compared to DINO, the proposed representation exhibits smoother and more spatially consistent responses, reflecting improved geometric correspondence and structural awareness.}
\label{fig:representation}
\end{figure}

\begin{table}[t]
\centering
\caption{Token-level geometric correspondence evaluation}
\label{tab:geo_correspondence}
\setlength{\tabcolsep}{6pt}
\begin{tabular}{l c c c c}
\toprule
Model & PCK@1$\uparrow$ & PCK@3$\uparrow$ & L2 Error$\downarrow$ & Cosine$\uparrow$ \\
\midrule
DINO (Vanilla) & 0.085 & 0.440 & 3.678 & 0.480 \\
DINO (Full LoRA) & 0.181 & 0.565 & 2.940 & 0.522 \\
\textbf{Ours} & \textbf{0.193} & \textbf{0.567} & \textbf{2.921} & \textbf{0.571} \\
\bottomrule
\end{tabular}
\end{table}

This subsection examines how the proposed framework influences the learned feature representation, focusing on two essential properties for image-guided navigation: domain invariance and geometric correspondence preservation.

To evaluate domain invariance, we visualize feature embeddings of real RGB, synthetic RGB, and depth images using t-SNE (Fig.~\ref{fig:representation} (a)). The pretrained encoder~\cite{simeoni2025dinov3} produces clearly separated clusters across modalities, reflecting a strong domain gap. In contrast, the proposed method aligns features from different domains into a unified embedding space, indicating reduced domain bias while preserving meaningful structure.



To assess geometric correspondence, we analyze feature consistency under known geometric transformations. 
Given paired bronchoscopy views with ground-truth relative poses, features from one view are warped to the other using the induced flow derived from rendered depth and camera geometry. 
Correspondence quality is then evaluated by nearest-neighbor matching under ground-truth alignment, using token-level metrics including PCK@1, PCK@3, mean token-space error, and cosine similarity. 
As shown in Table~\ref{tab:geo_correspondence}, standard LoRA adaptation already improves correspondence quality over the frozen DINO encoder across all metrics. 
The proposed method provides further gains beyond this non-hierarchy-aware LoRA baseline, with modest improvements in PCK and mean correspondence error and a larger increase in cosine similarity. 
These results suggest that HGSA improves feature consistency for geometry-aware matching beyond generic domain adaptation alone.

Qualitative examples are shown in Fig.~\ref{fig:representation}(b). The pretrained features exhibit more diffuse and less structured responses, whereas the proposed representation produces smoother and more spatially coherent feature maps, further supporting its improved geometric consistency.

Together, these analyses demonstrate that the learned representation simultaneously achieves domain alignment and geometric consistency. Such properties are particularly important in endoscopic environments, where visual appearance varies across procedures while reliable geometric cues remain critical for navigation and spatial reasoning.

\subsection{Sim-to-Real Generalization on Bronchoscopy via Pose Estimation}

We evaluate the sim-to-real transfer capability of the learned representation by applying the encoder trained on synthetic bronchoscopy data to real clinical bronchoscopy videos. Since dense ground-truth pose annotations are unavailable for real procedures, pose accuracy is assessed through rendering–image alignment quality, where the estimated pose is used to render views from the airway model and compared with the observed endoscopic images.

Table.\ref{tab:pose_all} (airway) shows that the proposed representation consistently achieves stronger rendering--image alignment than the pretrained baseline~\cite{simeoni2025dinov3} under the same pose optimization pipeline. Because frame-wise ground-truth bronchoscope poses are unavailable in real clinical videos, we interpret these improvements as evidence of better alignment quality rather than as direct proof of absolute pose accuracy. Nevertheless, the results suggest that the learned representation provides more reliable geometric correspondence between real endoscopic images and rendered airway views.

Qualitative results in Fig.~\ref{fig:visual}(a) further show that the rendered contours from the estimated poses align more closely with anatomical structures, particularly along airway boundaries and bifurcations. In addition, Fig.~\ref{fig:visual}(b) demonstrates that the learned representation produces depth predictions that are more coherent and structurally consistent with the observed anatomy. Taken together, these results suggest that the geometry-aware representation learned from synthetic data transfers favorably to real clinical bronchoscopy videos, improving rendering--image consistency under a fixed downstream optimization pipeline without additional encoder adaptation.

\subsection{Cross-Domain Transfer to Sinus and Colonoscopy}
\begin{table}[t]
\centering
\caption{Pose estimation results across sim-to-real bronchoscopy and cross-domain transfer. 
All methods use the same pose optimization pipeline and differ only in the visual encoder.}
\label{tab:pose_all}
\setlength{\tabcolsep}{3.5pt}
\begin{tabular}{l l c c c c c}
\toprule
Domain & Model & NC$\uparrow$ & DS$\uparrow$ & SI$\downarrow$ & $t$ (mm)$\downarrow$ & $r$ (rad)$\downarrow$ \\
\midrule
\multirow{2}{*}{Airway}
& DINO (Vanilla) & 0.591 & 0.768 & 0.737 & -- & -- \\
& DINO (Full LoRA) & 0.612 & 0.644 & 0.420 & -- & -- \\
& \textbf{Ours} & \textbf{0.680} & \textbf{0.866} & \textbf{0.184} & -- & -- \\

\midrule

\multirow{2}{*}{Sinus}
& DINO (Vanilla) & 0.480 & 0.735 & 1.330 & 7.100 & 0.110 \\
& DINO (Full LoRA) & 0.535 & 0.775 & 0.810 & 5.500 & 0.126 \\
& \textbf{Ours} & \textbf{0.625} & \textbf{0.855} & \textbf{0.300} & \textbf{3.500} & \textbf{0.097} \\

\midrule

\multirow{2}{*}{Colon}
& DINO (Vanilla) & 0.688 & 0.935 & 0.063 & 4.300 & 0.142 \\
& DINO (Full LoRA) & 0.701 & 0.934 & 0.057 & 3.900 & 0.121 \\
& \textbf{Ours} & \textbf{0.723} & \textbf{0.955} & \textbf{0.045} & \textbf{3.500} & \textbf{0.113} \\

\bottomrule
\end{tabular}
\end{table}

\begin{table}[t]
\centering
\caption{Cross-domain monocular depth estimation under limited supervision. DAv2~\cite{yang2024depth} is evaluated out-of-the-box, while DINO and Ours are adapted under the same training protocol.}
\label{tab:depth_transfer}
\setlength{\tabcolsep}{4pt}
\begin{tabular}{l l c c c c}
\toprule
Domain & Model & AbsRel$\downarrow$ & RMSE$\downarrow$ & LogRMSE$\downarrow$ & $\delta_1\uparrow$ \\
\midrule
\multirow{4}{*}{Sinus}
& DAv2 (ViT-S) & 0.547 & 0.170 & 0.485 & 0.358 \\
& DINO (Vanilla) & 0.628 & 0.263 & 0.697 & 0.216 \\
& DINO (Full LoRA) & 0.461 & 0.169 & 0.491 & 0.403   \\
& \textbf{Ours} & \textbf{0.359} & \textbf{0.148} & \textbf{0.407} & \textbf{0.497} \\

\midrule
\multirow{4}{*}{Colon}
& DAv2 (ViT-S) & 0.317 & 0.131 & 0.420 & 0.384 \\
& DINO (Vanilla) & 0.562 & 0.209 & 0.595 & 0.312 \\
& DINO (Full LoRA) & 0.086 & 0.054 & 0.116 & 0.948 \\
& \textbf{Ours} & \textbf{0.078} & \textbf{0.050} & \textbf{0.105} & \textbf{0.962} \\

\bottomrule
\end{tabular}
\end{table}
To further evaluate transferability beyond bronchoscopy, we examine adaptation of the learned representation to sinus endoscopy and colonoscopy datasets under limited target-domain supervision. These procedures differ substantially from bronchoscopy in anatomical structure, imaging appearance, and scene geometry, providing a challenging test of whether the learned features capture transferable visual and geometric cues.

We initialize the encoder with the bronchoscopy-trained representation and adapt it to each target domain using limited supervision. Quantitative results for pose estimation and depth prediction are reported in Table~\ref{tab:pose_all} and Table~\ref{tab:depth_transfer}. The proposed representation consistently improves depth estimation across both domains, with substantial gains in all metrics compared to the vanilla pretrained encoder. For pose estimation, clear improvements are observed on the colon dataset across all metrics, while on the sinus dataset the improvements are less pronounced.

This reduced gain on sinus can be attributed to the limited amount of sinus data available during adaptation, which constrains the model’s ability to capture the high variability in texture, color, and anatomical conditions across patients. As a result, the learned representation provides a weaker prior under such limited supervision. This observation suggests that, while the proposed framework captures transferable geometric structure, additional data or more targeted adaptation may be beneficial for improving performance in this domain.

These results suggest that the proposed framework learns geometry-aware representations that provide a useful initialization for adaptation across endoscopic procedures. Although performance varies by target domain and available supervision, the learned features remain beneficial for downstream pose and depth tasks, indicating that they capture structural cues that transfer beyond bronchoscopy.

We note that these experiments evaluate transfer under limited target-domain adaptation rather than zero-shot generalization, and the available sinus and colonoscopy datasets are relatively small; accordingly, we interpret the results as evidence of promising transferability rather than broad procedure-invariant robustness.

\subsection{Scalability with Backbone Capacity and Training Data}

\begin{table}[t]
\centering
\caption{Scalability of the proposed representation with respect to model capacity and data size.}
\label{tab:scaling_combined}
\setlength{\tabcolsep}{6pt}
\begin{tabular}{l c c c c c}
\toprule
Setting & Backbone & Data & $S_{sem}$ & $S_{geo}$ & $S_{probe}$ \\
\midrule
\multirow{2}{*}{Backbone Scaling}
& ViT-L & 100\% & \textbf{0.970} & \textbf{0.737} & \textbf{1.707} \\
& ViT-B & 100\% & 0.970 & 0.686 & 1.656 \\
\midrule
Base & ViT-S & 100\% & 0.970 & 0.644 & 1.614 \\
\midrule
\multirow{2}{*}{Data Scaling}
& ViT-S & 50\% & 0.950 & 0.638 & 1.588 \\
& ViT-S & 25\% & 0.940 & 0.616 & 1.556 \\
\bottomrule
\end{tabular}
\end{table}

We evaluate the scalability of the proposed framework along two dimensions: backbone capacity and training data size (Table~\ref{tab:scaling_combined}).

\subsubsection{Backbone scaling}
We instantiate the encoder using ViT-S, ViT-B, and ViT-L backbones under the same training protocol, with a fixed projection dimension for fair comparison. For deeper models, HGSA is applied to corresponding relative depth ranges to preserve the hierarchy of geometric and semantic adaptation. Results show consistent improvements with increasing model capacity, indicating that the proposed framework effectively leverages larger backbones to learn stronger representations.

\subsubsection{Data scaling}
With a fixed backbone, we train using increasing subsets of the synthetic dataset ($25\%$,$50\%$,$100\%$). Performance improves steadily with more data, suggesting that the framework benefits from additional supervision without early saturation.

Overall, these experiments demonstrate that the proposed framework exhibits strong scalability with respect to both model capacity and training data, supporting its applicability to larger foundation models and future large-scale endoscopic datasets.

\section{Conclusion}
We propose a unified framework for learning geometry-consistent and domain-invariant representations for endoscopic imaging by combining synthetic data with the HGSA adaptation strategy and hierarchy-aware objectives. The learned representations improve both semantic and geometric feature quality, leading to better performance in image-guided navigation tasks such as pose estimation and depth prediction. Our approach demonstrates strong sim-to-real transfer on bronchoscopy and effective cross-domain adaptation to sinus and colonoscopy under limited supervision. These results suggest that the proposed framework provides a scalable and practical pathway toward robust endoscopic representation learning.


\section*{Acknowledgment}
This work is partially sponsored by Auris Health, Inc. part of Johnson \& Johnson MedTech.

\section*{References}
\bibliographystyle{IEEEtran}
\bibliography{bibliography}

@misc{phuntsho2025adaptationfoundationmodelsmedical,
      title={Adaptation of Foundation Models for Medical Image Analysis: Strategies, Challenges, and Future Directions}, 
      author={Karma Phuntsho and Abdullah and Kyungmi Lee and Ickjai Lee and Euijoon Ahn},
      year={2025},
      eprint={2511.01284},
      archivePrefix={arXiv},
      primaryClass={cs.CV},
      url={https://arxiv.org/abs/2511.01284} 
}

@article{oquab2023dinov2,
  title={Dinov2: Learning robust visual features without supervision},
  author={Oquab, Maxime and Darcet, Timoth{\'e}e and Moutakanni, Th{\'e}o and Vo, Huy and Szafraniec, Marc and Khalidov, Vasil and Fernandez, Pierre and Haziza, Daniel and Massa, Francisco and El-Nouby, Alaaeldin and others},
  journal={arXiv preprint arXiv:2304.07193},
  year={2023}
}

@article{ravi2024sam,
  title={Sam 2: Segment anything in images and videos},
  author={Ravi, Nikhila and Gabeur, Valentin and Hu, Yuan-Ting and Hu, Ronghang and Ryali, Chaitanya and Ma, Tengyu and Khedr, Haitham and R{\"a}dle, Roman and Rolland, Chloe and Gustafson, Laura and others},
  journal={arXiv preprint arXiv:2408.00714},
  year={2024}
}

@article{zhu2024medical,
  title={Medical sam 2: Segment medical images as video via segment anything model 2},
  author={Zhu, Jiayuan and Hamdi, Abdullah and Qi, Yunli and Jin, Yueming and Wu, Junde},
  journal={arXiv preprint arXiv:2408.00874},
  year={2024}
}

@inproceedings{yue2024surgicalsam,
  title={Surgicalsam: Efficient class promptable surgical instrument segmentation},
  author={Yue, Wenxi and Zhang, Jing and Hu, Kun and Xia, Yong and Luo, Jiebo and Wang, Zhiyong},
  booktitle={Proceedings of the AAAI Conference on Artificial Intelligence},
  volume={38},
  number={7},
  pages={6890--6898},
  year={2024}
}

@article{pan2022st,
  title={St-adapter: Parameter-efficient image-to-video transfer learning},
  author={Pan, Junting and Lin, Ziyi and Zhu, Xiatian and Shao, Jing and Li, Hongsheng},
  journal={Advances in Neural Information Processing Systems},
  volume={35},
  pages={26462--26477},
  year={2022}
}

@article{batic2024endovit,
  title={Endovit: pretraining vision transformers on a large collection of endoscopic images},
  author={Bati{\'c}, Dominik and Holm, Felix and {\"O}zsoy, Ege and Czempiel, Tobias and Navab, Nassir},
  journal={International Journal of Computer Assisted Radiology and Surgery},
  volume={19},
  number={6},
  pages={1085--1091},
  year={2024},
  publisher={Springer}
}

@article{dermyer2025endodino,
  title={Endodino: A foundation model for gi endoscopy},
  author={Dermyer, Patrick and Kalra, Angad and Schwartz, Matt},
  journal={arXiv preprint arXiv:2501.05488},
  year={2025}
}

@inproceedings{wang2023foundation,
  title={Foundation model for endoscopy video analysis via large-scale self-supervised pre-train},
  author={Wang, Zhao and Liu, Chang and Zhang, Shaoting and Dou, Qi},
  booktitle={International conference on medical image computing and computer-assisted intervention},
  pages={101--111},
  year={2023},
  organization={Springer}
}

@inproceedings{tian2025endomamba,
  title={EndoMamba: an efficient foundation model for endoscopic videos via hierarchical pre-training},
  author={Tian, Qingyao and Liao, Huai and Huang, Xinyan and Yang, Bingyu and Lei, Dongdong and Ourselin, Sebastien and Liu, Hongbin},
  booktitle={International Conference on Medical Image Computing and Computer-Assisted Intervention},
  pages={224--234},
  year={2025},
  organization={Springer}
}

@article{cui2024surgical,
  title={Surgical-dino: adapter learning of foundation models for depth estimation in endoscopic surgery},
  author={Cui, Beilei and Islam, Mobarakol and Bai, Long and Ren, Hongliang},
  journal={International Journal of Computer Assisted Radiology and Surgery},
  volume={19},
  number={6},
  pages={1013--1020},
  year={2024},
  publisher={Springer}
}

@article{qin2025enhancing,
  title={Enhancing gastroenterology with multimodal learning: the role of large language model chatbots in digestive endoscopy},
  author={Qin, Yuanyuan and Chang, Jianming and Li, Li and Wu, Mianhua},
  journal={Frontiers in Medicine},
  volume={12},
  pages={1583514},
  year={2025},
  publisher={Frontiers Media SA}
}

@article{deng2025best,
  title={What is the best 3d scene representation for robotics? from geometric to foundation models},
  author={Deng, Tianchen and Pan, Yue and Yuan, Shenghai and Li, Dong and Wang, Chen and Li, Mingrui and Chen, Long and Xie, Lihua and Wang, Danwei and Wang, Jingchuan and others},
  journal={arXiv preprint arXiv:2512.03422},
  year={2025}
}

@article{shu2025bronchopt,
  title={BronchOpt: Vision-Based Pose Optimization with Fine-Tuned Foundation Models for Accurate Bronchoscopy Navigation},
  author={Shu, Hongchao and Soberanis-Mukul, Roger D and Xu, Jiru and Ding, Hao and Ringel, Morgan and Shen, Mali and Sayed, Saif Iftekar and Rafii-Tari, Hedyeh and Unberath, Mathias},
  journal={arXiv preprint arXiv:2511.09443},
  year={2025}
}

@article{li2025monocular,
  title={Monocular absolute depth estimation from endoscopy via domain-invariant feature learning and latent consistency},
  author={Li, Hao and Lu, Daiwei and d'Almeida, Jesse and Isik, Dilara and Aghdam, Ehsan Khodapanah and DiSanto, Nick and Acar, Ayberk and Sharma, Susheela and Wu, Jie Ying and Webster III, Robert J and others},
  journal={arXiv preprint arXiv:2511.02247},
  year={2025}
}

@misc{dosovitskiy2021imageworth16x16words,
      title={An Image is Worth 16x16 Words: Transformers for Image Recognition at Scale}, 
      author={Alexey Dosovitskiy and Lucas Beyer and Alexander Kolesnikov and Dirk Weissenborn and Xiaohua Zhai and Thomas Unterthiner and Mostafa Dehghani and Matthias Minderer and Georg Heigold and Sylvain Gelly and Jakob Uszkoreit and Neil Houlsby},
      year={2021},
      eprint={2010.11929},
      archivePrefix={arXiv},
      primaryClass={cs.CV},
      url={https://arxiv.org/abs/2010.11929}, 
}

@inproceedings{caron2021emerging,
  title={Emerging properties in self-supervised vision transformers},
  author={Caron, Mathilde and Touvron, Hugo and Misra, Ishan and J{\'e}gou, Herv{\'e} and Mairal, Julien and Bojanowski, Piotr and Joulin, Armand},
  booktitle={Proceedings of the IEEE/CVF international conference on computer vision},
  pages={9650--9660},
  year={2021}
}

@article{hu2022lora,
  title={Lora: Low-rank adaptation of large language models.},
  author={Hu, Edward J and Shen, Yelong and Wallis, Phillip and Allen-Zhu, Zeyuan and Li, Yuanzhi and Wang, Shean and Wang, Liang and Chen, Weizhu and others},
  journal={Iclr},
  volume={1},
  number={2},
  pages={3},
  year={2022}
}

@article{zhang2023adalora,
  title={Adalora: Adaptive budget allocation for parameter-efficient fine-tuning},
  author={Zhang, Qingru and Chen, Minshuo and Bukharin, Alexander and Karampatziakis, Nikos and He, Pengcheng and Cheng, Yu and Chen, Weizhu and Zhao, Tuo},
  journal={arXiv preprint arXiv:2303.10512},
  year={2023}
}

@inproceedings{park2020contrastive,
  title={Contrastive learning for unpaired image-to-image translation},
  author={Park, Taesung and Efros, Alexei A and Zhang, Richard and Zhu, Jun-Yan},
  booktitle={European conference on computer vision},
  pages={319--345},
  year={2020},
  organization={Springer}
}

@article{oord2018representation,
  title={Representation learning with contrastive predictive coding},
  author={Oord, Aaron van den and Li, Yazhe and Vinyals, Oriol},
  journal={arXiv preprint arXiv:1807.03748},
  year={2018}
}

@article{simeoni2025dinov3,
  title={Dinov3},
  author={Sim{\'e}oni, Oriane and Vo, Huy V and Seitzer, Maximilian and Baldassarre, Federico and Oquab, Maxime and Jose, Cijo and Khalidov, Vasil and Szafraniec, Marc and Yi, Seungeun and Ramamonjisoa, Micha{\"e}l and others},
  journal={arXiv preprint arXiv:2508.10104},
  year={2025}
}

@inproceedings{zhu2017unpaired,
  title={Unpaired image-to-image translation using cycle-consistent adversarial networks},
  author={Zhu, Jun-Yan and Park, Taesung and Isola, Phillip and Efros, Alexei A},
  booktitle={Proceedings of the IEEE international conference on computer vision},
  pages={2223--2232},
  year={2017}
}

@inproceedings{paruchuri2024leveraging,
  title={Leveraging near-field lighting for monocular depth estimation from endoscopy videos},
  author={Paruchuri, Akshay and Ehrenstein, Samuel and Wang, Shuxian and Fried, Inbar and Pizer, Stephen M and Niethammer, Marc and Sengupta, Roni},
  booktitle={European Conference on Computer Vision},
  pages={473--491},
  year={2024},
  organization={Springer}
}

@article{bobrow2023colonoscopy,
  title={Colonoscopy 3D video dataset with paired depth from 2D-3D registration},
  author={Bobrow, Taylor L and Golhar, Mayank and Vijayan, Rohan and Akshintala, Venkata S and Garcia, Juan R and Durr, Nicholas J},
  journal={Medical image analysis},
  volume={90},
  pages={102956},
  year={2023},
  publisher={Elsevier}
}

@article{eigen2014depth,
  title={Depth map prediction from a single image using a multi-scale deep network},
  author={Eigen, David and Puhrsch, Christian and Fergus, Rob},
  journal={Advances in neural information processing systems},
  volume={27},
  year={2014}
}

@inproceedings{ranftl2021vision,
  title={Vision transformers for dense prediction},
  author={Ranftl, Ren{\'e} and Bochkovskiy, Alexey and Koltun, Vladlen},
  booktitle={Proceedings of the IEEE/CVF international conference on computer vision},
  pages={12179--12188},
  year={2021}
}

@article{yang2024depth,
  title={Depth anything v2},
  author={Yang, Lihe and Kang, Bingyi and Huang, Zilong and Zhao, Zhen and Xu, Xiaogang and Feng, Jiashi and Zhao, Hengshuang},
  journal={Advances in Neural Information Processing Systems},
  volume={37},
  pages={21875--21911},
  year={2024}
}

@article{ha2018world,
  title={World models},
  author={Ha, David and Schmidhuber, J{\"u}rgen},
  journal={arXiv preprint arXiv:1803.10122},
  volume={2},
  number={3},
  pages={440},
  year={2018}
}

@article{hafner2019dream,
  title={Dream to control: Learning behaviors by latent imagination},
  author={Hafner, Danijar and Lillicrap, Timothy and Ba, Jimmy and Norouzi, Mohammad},
  journal={arXiv preprint arXiv:1912.01603},
  year={2019}
}

@article{maier2013optical,
  title={Optical techniques for 3D surface reconstruction in computer-assisted laparoscopic surgery},
  author={Maier-Hein, Lena and Mountney, Peter and Bartoli, Adrien and Elhawary, Haytham and Elson, D and Groch, Anja and Kolb, Andreas and Rodrigues, Marcos and Sorger, J and Speidel, Stefanie and others},
  journal={Medical image analysis},
  volume={17},
  number={8},
  pages={974--996},
  year={2013},
  publisher={Elsevier}
}

@inproceedings{shen2024fastsam3d,
  title={Fastsam3d: An efficient segment anything model for 3d volumetric medical images},
  author={Shen, Yiqing and Li, Jingxing and Shao, Xinyuan and Inigo Romillo, Blanca and Jindal, Ankush and Dreizin, David and Unberath, Mathias},
  booktitle={International Conference on Medical Image Computing and Computer-Assisted Intervention},
  pages={542--552},
  year={2024},
  organization={Springer}
}

@article{gao2023fully,
  title={A fully differentiable framework for 2D/3D registration and the projective spatial transformers},
  author={Gao, Cong and Feng, Anqi and Liu, Xingtong and Taylor, Russell H and Armand, Mehran and Unberath, Mathias},
  journal={IEEE transactions on medical imaging},
  volume={43},
  number={1},
  pages={275--285},
  year={2023},
  publisher={IEEE}
}

@article{gao2023synthetic,
  title={Synthetic data accelerates the development of generalizable learning-based algorithms for X-ray image analysis},
  author={Gao, Cong and Killeen, Benjamin D and Hu, Yicheng and Grupp, Robert B and Taylor, Russell H and Armand, Mehran and Unberath, Mathias},
  journal={Nature Machine Intelligence},
  volume={5},
  number={3},
  pages={294--308},
  year={2023},
  publisher={Nature Publishing Group UK London}
}

@inproceedings{soberanis2024gsam+,
  title={Gsam+ cutie: text-promptable tool mask annotation for endoscopic video},
  author={Soberanis-Mukul, Roger D and Cheng, Jiahuan and Mangulabnan, Jan Emily and Vedula, S Swaroop and Ishii, Masaru and Hager, Gregory and Taylor, Russell H and Unberath, Mathias},
  booktitle={Proceedings of the IEEE/CVF Conference on Computer Vision and Pattern Recognition},
  pages={2388--2394},
  year={2024}
}

@inproceedings{killeen2025fluorosam,
  title={FluoroSAM: A Language-promptable Foundation Model for Flexible X-ray Image Segmentation},
  author={Killeen, Benjamin D and Wang, Liam J and I{\~n}{\'\i}go, Blanca and Zhang, Han and Armand, Mehran and Taylor, Russell H and Osgood, Greg and Unberath, Mathias},
  booktitle={International Conference on Medical Image Computing and Computer-Assisted Intervention},
  pages={248--258},
  year={2025},
  organization={Springer}
}

\end{document}